\documentclass[11pt]{article}
\usepackage{coling2018}
\usepackage{times}
\usepackage{url}
\usepackage{latexsym}

\usepackage{nicefrac}       
\usepackage{xfrac}
\usepackage{microtype}      
\usepackage{mathtools}
\usepackage{changepage}

\usepackage{xspace}
\usepackage{caption}
\usepackage{wrapfig}

\usepackage{color}
\usepackage{multirow}

\usepackage{graphicx}

\usepackage{xcolor}
\usepackage{lipsum}

\usepackage{anyfontsize}

\makeatletter
\DeclareRobustCommand\onedot{\futurelet\@let@token\@onedot}
\def\@onedot{\ifx\@let@token.\else.\null\fi\xspace}

\def\eg{\emph{e.g}\onedot} 
\def\ie{\emph{i.e}\onedot}

\def\wrt{w.r.t\onedot} 

\makeatother

\title{Learning Visually-Grounded Semantics\\ from Contrastive Adversarial Samples\vspace{-2em}}

\author{
Haoyue Shi~${}^{1}$\thanks{\hspace{0.65em} Work was done when HS, JM and TX were intern researchers at Megvii Inc. HS, JM and TX contribute equally to this paper.}
\quad Jiayuan Mao~${}^{2*}$
\quad Tete Xiao~${}^{1*}$
\quad Yuning Jiang~${}^{3}$
\quad Jian Sun~${}^{3}$\\
}

\date{}

\begin{document}
\maketitle
{
\vspace{-7.5em}
\centering
${}^{1}$: School of Electronics Engineering and Computer Science, Peking University, China\\
${}^{2}$: ITCS, Institute for Interdisciplinary Information Sciences, Tsinghua University, China\\
${}^{3}$: Megvii, Inc.\\
{\tt \{hyshi,jasonhsiao97\}@pku.edu.cn}\\
{\tt mjy14@mails.tsinghua.edu.cn, \{jyn, sunjian\}@megvii.com}

\vspace{1em}
}

\begin{abstract}
We study the problem of grounding distributional representations of texts on the visual domain, namely visual-semantic embeddings (VSE for short).
Begin with an insightful adversarial attack on VSE embeddings, we show the limitation of current frameworks and image-text datasets (\eg, MS-COCO) both quantitatively and qualitatively. The large gap between the number of possible constitutions of real-world semantics and the size of parallel data, to a large extent, restricts the model to establish the link between textual semantics and visual concepts. We alleviate this problem by augmenting the MS-COCO image captioning datasets with textual contrastive adversarial samples. These samples are synthesized using linguistic rules and the WordNet knowledge base. The construction procedure is both syntax- and semantics-aware. The samples enforce the model to ground learned embeddings to concrete concepts within the image. This simple but powerful technique brings a noticeable improvement over the baselines on a diverse set of downstream tasks, in addition to defending known-type adversarial attacks. We release the codes at \url{https://github.com/ExplorerFreda/VSE-C}.
\end{abstract}

%
%
\blfootnote{
    %
    %
    \hspace{-0.65cm}  
    %
    %
    %
    %
    This work is licensed under a Creative Commons 
    Attribution 4.0 International License.
    License details:
    \url{http://creativecommons.org/licenses/by/4.0/}
}

\section{Introduction}
\label{sec:introduction}

The visual grounding of language plays an indispensable role in our daily lives. We use language to name, refer, and describe objects, their properties and generally, visual concepts.
Distributional semantics (\eg, global word embeddings~\cite{pennington2014glove}) based on large-scale corpora have shown great success in modeling the functionality and correlation of words in the natural language domain. This further contributes to the success in numerous natural language processing (NLP) tasks such as language modeling~\cite{cheng2016long,inan2016tying}, sentiment analysis~\cite{cheng2016long,kumar2016ask}, and reading comprehension ~\cite{cheng2016long,chen2016thorough,shen2017reasonet}. However, effective and efficient grounding of distributional embeddings remains challenging.
Being ignorant of the corresponding visual concepts, pure textual embeddings demonstrate inferior performances when incorporating with visual inputs.
A set of typical tasks includes image/video captioning, multi-modal retrieval/understanding, and visual reasoning, some of which are further extensively studied in the paper.


Visual concept and its link with textual semantics, as a cognitive alignment, provide rich supervision to learning systems. Introduced in \newcite{kiros2014unifying}, Visual-Semantic Embedding (VSE) aims at building the bridge between natural language and the underlying visual world by jointly optimize and align the embedding spaces of both images and descriptive texts (captions). Nevertheless, even for large-scale datasets such as MS-COCO~\cite{lin2014microsoft}, the number of image-caption pairs are far less than the number of possible constitutions of real-world semantics, making the dataset inevitably sparse and biased.

\begin{figure}[!tb]
\includegraphics[width=\textwidth]{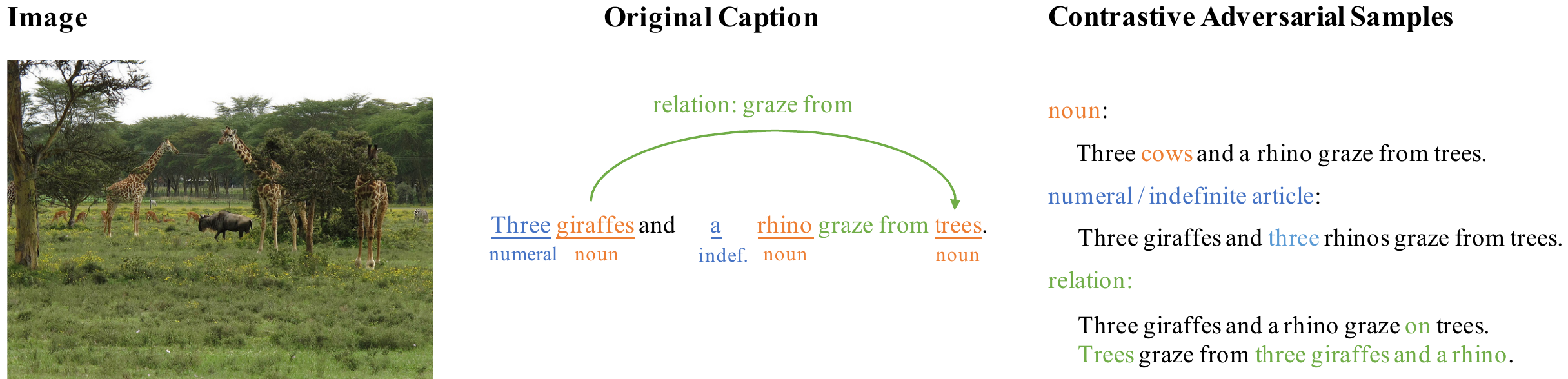}
\caption{An overview of our textual contrastive adversarial samples. For each caption, we have three paradigms to generate contrastive adversarial samples, \textit{i.e.}, noun, numeral and relation. For an given image, we expect the model to distinguish the real captions against the generated adversarial ones.}
\label{fig:intro}
\vspace{-1em}
\end{figure}

To reveal this, we begin with constructing textual adversarial samples to attack the state-of-the-art system VSE++~\cite{faghri2017vse++}. Specifically, we study the composition of sentences from two aspects: (1) content words including nouns and numerals and (2) prepositions indicating spatial relations (\eg, in, on, above, below). As shown in Figure~\ref{fig:intro}, we manipulate the original caption to construct hard negative captions with similar structure but completely contradictory semantics. We found that the models easily get confused, suffering a noticeable drop in confidence or even wrong predictions in the caption retrieval task.

We propose VSE-C, which enforces the learning of correlation and correspondence between textual semantics and visual concepts by providing contrastive adversarial samples during the training procedure, incorporating with an intra-pair hard negative sample mining.
Instead of defending adversarial attacks, we focus on the study of limitations of current visual-semantic datasets and the transferability of learned embeddings.
To fulfill the large gap between the number of parallel image-caption pairs and the expressiveness of natural languages, we augment the data by employing a set of heuristic rules to generate large sets of contrastive negative captions, as demonstrated in Figure~\ref{fig:intro}. The candidates are selectively used for training by an intra-pair hard-example mining technique. VSE-C alleviate the bias of dataset and provide rich and effective samples on par with original image captions. This strengthen the link between text and visual concepts by requiring models to detect a mismatch on the level of some precise concepts.

VSE-C learns discriminative and visually-grounded word embeddings on the MS-COCO dataset~\cite{lin2014microsoft}. It is extensively compared with existing works with rich experiments and analyses. Most importantly, we explore the transferability of the learned embeddings on several real-world applications both qualitatively and quantitatively, including image-to-text retrieval and bidirectional word-to-concept retrieval. Furthermore, VSE-C demonstrates a general framework for augmenting textual inputs considering semantical consistency. The introduction human priors and knowledge bases alleviates the sparsity and non-contiguity of languages.
We release our codes and data at \url{https://github.com/ExplorerFreda/VSE-C}.

\section{Related works}
\paragraph{Joint embeddings}
Joint embedding is a common technique for a wide range of tasks incorporating multiple domains, including audio-video embeddings for unsupervised representation learning~\cite{ngiam2011multimodal}, shape-image embeddings~\cite{li2015joint} for shape inference, bilingual word embeddings for machine translation~\cite{zou2013bilingual}, human pose-image embeddings for pose inference~\cite{li2011learning}, image-text embeddings for visual description~\cite{reed2016learning}, and global representation learning from multiple domains~\cite{castrejon2016learning}. These embeddings map multiple domains into a joint vector space which describes the semantical relations between inputs (\eg, distance, correlation).

We focus on the visual-semantic embedding~\cite{mao2016training,kiros2014unifying,faghri2017vse++}, learning word embeddings with visually-grounded semantics. Examples of related applications include image caption retrieval and generation~\cite{kiros2014unifying,karpathy2015deep}, and visual question-answering~\cite{malinowski2015ask}.

\paragraph{Image-to-text translation}
Canonical Correlation Analysis (CCA)~\cite{hotelling1936relations} is a statistical method that projects two views linearly into a common space to maximize their correlation. \newcite{andrew2013deep} proposes a deep learning framework to extend CCA so that it is able to learn nonlinear projections and has better scalability on relatively large datasets. 

In the state-of-the-art frameworks, the pairwise ranking is often adopted to learn a distance metric~\cite{socher2014grounded,niu2017hierarchical,nam2017dual}.
\newcite{frome2013devise} proposes a cross-modal feature embedding framework that uses CNN and Skip-Gram~\cite{mikolov2013efficient} to extract representations for images and texts respectively, then an objective is applied to ensure that the distance between the matched image-text pair is smaller than that between the mismatched pair.
A similar framework proposed by \newcite{kiros2014unifying} uses a Gated Recurrent Unit (GRU) as the sentence encoder. 
\newcite{wang2016learning} uses a bidirectional loss function with structure-preserving constraints. 
An attention mechanism on both image and caption is used by \newcite{nam2017dual} where the model estimates the similarity between images and texts by sequentially focusing on a subset of image regions and words that have shared semantics. 
\newcite{huang2017instance} utilizes a multi-modal context-modulated attention mechanism to compute the similarity between an image and a caption. 
\newcite{faghri2017vse++} proposes a novel loss to penalize the hard negatives, \ie, the closest mismatched pairs, instead of averaging the individual violations across all negatives in \newcite{kiros2014unifying}.

\paragraph{Adversarial attack in text domain}
Adversarial attacks have recently drawn significant attention in the deep learning community. The adversarial attacks spread over multiple domains including image classification~\cite{nguyen2015deep}, image segmentation and object detection~\cite{xie2017adversarial}, and textual reading comprehension~\cite{jia2017adversarial}, and deep reinforcement learning~\cite{kos2017delving}.

In this paper, we present textual adversarial attacks in image-to-text translation systems such as image caption frameworks. While focusing on the problem of learning visually-grounded semantics, the adversarial attack brings new solutions to fulfill the gap between limited training data and numerous constitutions of natural languages.
With extensive experiments on the effects of the adversarial samples, we reach the conclusion that current visual-semantic embeddings are ``insensitive'' to the underlying semantics. The proposed VSE-C shows advance across multiple visual-semantic tasks.

\section{Method}
\subsection{Preliminaries}
\paragraph{Word embeddings}
We manually split the embeddings of each word into two parts: distributional embeddings, and visually-grounded embeddings. We use GloVe~\cite{pennington2014glove} as the distributional embeddings, pre-trained unsupervisedly on large-scale corpora. We focus on the visually-grounded embeddings of words. The embeddings are optimized using the visual-semantic embedding (VSE) technique.

\paragraph{Visual-semantic embeddings}
VSE optimizes and aligns the latent space of both visual and textual domains. Parallel data are typically obtained from image captioning datasets such as Flickr30K~\cite{young2014image} or MS-COCO~\cite{lin2014microsoft}. The training set $S = \{(i_n, c_n)\}_N$ contains $N$ image-caption pairs. Typically all $(i_n, c_m), n \neq m$ and $(i_m, c_n), n \neq m$ form the negative samples for a specific pair $(i_n, c_n)$.

Following the notations used by \newcite{kiros2014unifying}, domain-specific encoders are first employed to extract latent features of both images and captions, denoted as $\phi(i)$ and $\psi(c)$, respectively. We use ResNet-152~\cite{he2016deep} as visual domain encoder and GRU as text domain sentence encoder, which are both effective for VSE. They are projected into a joint latent space with a linear transformation. A hinge loss with margin $\alpha$ is employed to optimize the alignment:
\begin{equation}
\ell^{VSE}(i, c) = \sum_{c'} [\alpha + s(i, c') - s(i,c)]_+ + \sum_{i'} [\alpha + s(i',c) - s(i,c)]_+\,,
\end{equation}
where $[\cdot]_+ = \max(0, \cdot)$, and $s(i, c) = W_i^T f(i; \theta_i) \cdot W_c^T g(c; \theta_c)$ measuring the distance between projected image embedding $W_i^T f(i; \theta_i)$ and caption embedding $W_c^T g(c; \theta_c)$. The summations are taken over all image-caption pairs within a sampled batch.

\subsection{Generating contrastive adversarial samples}
\label{sec:method:manip}
\begin{table}[h]
\centering
\begin{tabular}{l|l|l}
\hline
\textbf{Class} & \textbf{Original Caption} &  \textbf{Contrastive Adversarial Example} \\
\hline
Noun & A person feeding a \underline{\textbf{cat}} with a banana. & A person feeding a \underline{\textbf{dog}} with a banana. \\
Numeral & A person feeding \underline{\textbf{a cat}} with a banana. & A person feeding \underline{\textbf{five cats}} with a banana.\\
Relation-1 & \underline{\textbf{A person}} feeding \underline{\textbf{a cat}} with a banana. & \underline{\textbf{A cat}} feeding \underline{\textbf{a person}} with a banana.\\
Relation-2 & A person feeding a cat \underline{\textbf{with}} a banana. & A person feeding a cat \underline{\textbf{in}} a banana.\\
\hline
\end{tabular}
\caption{Examples of contrastive adversarial samples generated with our heuristic rules and knowledge from WordNet. The samples can be classified into four types: noun replacement, numeral replacement, relation shuffling, and relation replacement.}
\label{table: contrastive-adversarial-examples}
\end{table}

Our contrastive adversarial samples can be split into three classes: \textit{noun}, \textit{numeral} and \textit{relation}. Each class of samples is generated separately. 
\paragraph{Noun.} We extract a list of heads~\cite{zwicky1985heads} of noun phrases in MS-COCO dataset and label those with frequency larger than 200 be frequent heads. In addition, since images usually reflect concrete concepts better than abstract ones, we compute the concreteness of words following \newcite{turney2011literal}, and only consider those heads with concreteness larger than $\theta=0.6$. Only frequent concrete heads can be replaced by other frequent concrete heads with different meaning to form contrastive adversarial samples.

While replacing, we utilize the hypernymy/hyponymy relations in WordNet~\cite{miller1995wordnet} to confirm the original noun and the corresponding contrastive adversarial sample are semantically different.
Only words without hypernymy or hyponymy relations can be used as the replacement for adversarial sample generation. 
For example, ``animal'' is a hypernym of ``cat''. Therefore, ``A person feeding an animal with a banana'' cannot be a valid generated contrastive adversarial caption for the image with the caption of ``A person feeding a cat with a banana.''

\paragraph{Numeral.} For each caption, we detect numerals and replace them with other numerals indicating different quantities to form contrastive adversarial samples. 
Note that ``a'' and ``an'' are treated as ``one'' here, though they are (indefinite) articles instead of numerals. 
Meanwhile, we singularize or pluralize the corresponding nouns when necessary. 
\paragraph{Relation.} The relation class includes two different paradigms. 

The first one can be viewed as \textit{shuffle of noninterchangeable noun phrases}. After extracting noun phrases of a caption, we shuffle them and put them back to the original positions. 
Although the bag of words features of the two sentences (caption) remain the same, the semantic meaning alters through this process.

The second one is \textit{replacement of prepositions.} We extract the prepositions with frequency higher than 200 in MS-COCO dataset. Then we manually annotate a semantic overlap table, which can be found in Appendix~\ref{appendix: relation}. In this table, words in the same set may have semantic overlap with each other, \textit{e.g.}, by and with, in and among. 

The noun phrase detection, preposition detection and numeral detection mentioned above are performed with SpaCy~\cite{honnibal2015improved}. Examples of different classes of contrastive adversarial sample generation are shown in Table~\ref{table: contrastive-adversarial-examples}. 

\subsection{Intra-pair hard negative mining}
We extend the online hard example mining (OHEM) technique used by VSE++~\cite{faghri2017vse++}. The original hinge loss is computed by choosing the hardest sample within an batch (inter-pair). Mathematically,
\begin{equation} 
\ell^{\text{VSE++}}(i, c) = \max_{c' \neq c} [\alpha + s(i, c') - s(i, c)] + \max_{i' \neq i} [\alpha + s(i', c) - s(i, c)]. 
\end{equation}

There are two major concerns regarding the in-batch hard negative mining. On one hand, mining negatives from a single batch is inefficient when batch size is not comparable with the size of the dataset. On the other hand, for real-world datasets, taking the max in loss function tends to be very sensitive to label noise, resulting in fake negative samples.

In contrast, given an image-caption pair $(i, c)$, we employ human heuristics and WordNet knowledge base to generate contrastive negative samples $\mathcal C'(c)$. To utilize these candidate caption sets, we employ an intra-pair hard negative mining strategy. Specifically, during the optimization, we add an extra loss term:

\begin{equation} 
\ell^{\text{VSE-C}}(i, c) = \ell^{\text{VSE++}}(i, c) + \max_{c'' \in \mathcal C'(c)} [\alpha + s(i, c'') - s(i, c)]_+. 
\end{equation}

In our implementation, the candidate set $\mathcal{C'}$ has approximately $1,000$ samples. In each iteration, we randomly sample $N=8$ negatives from it. This simple sample technique are effective and computation-friendly based on our empirical studies.

\section{Experiments}

We begin our experiments with an extensive study on the effect of adversarial samples on the baseline models. Even trained with hard negative mining techniques, VSE++ fails to discriminate words with completely contradictory visually-grounded semantics. Furthermore, we study the improvement brought by the introduction of contrastive adversarial samples on a diverse set of tasks. 

\subsection{Adversarial attacks}
\label{sec: adversarial-attack}

We select 1,000 images for test in MS-COCO 5k test split following \newcite{karpathy2015deep}. Each image is associated with five captions. Each caption in the selected test set can be manipulated to generate at least 20 contrastive adversarial samples by all manners (noun, numeral, and relation adversary). The image-to-caption retrieval task is defined as ranking the candidate captions based on the distance between their semantics and the given image.

We follow the metric used in \newcite{faghri2017vse++} computing R@1, R@10, median rank and mean rank \wrt the top-ranked correct caption for each image. For each image, the database of retrieval contains the full set of ~$1000\times{}5$ captions, in which only 5 captions are labeled as positive. The R@k metric essentially measures the percentage of images where the set of top-k ranked captions contains at least one positive caption.

We attack the existing models by adversarial samples. We extend each caption with 60 adversarial samples (20 noun-typed, 20 numeral-typed and 20 relation-typed). Therefore, each image has 60 $\times$ 5 contrastive adversarial samples in total. The candidate retrieval set for each image now becomes 5000 $+$ 300. We discuss the experimental results as follows:

\paragraph{VSE-C are more robust to known-typed adversarial attacks than VSE and VSE++.}
We compare the performance of VSE-C with VSE~\cite{kiros2014unifying} and VSE++~\cite{faghri2017vse++} in Table~\ref{tab:adversarial-attack}. Both VSE and VSE++ have a significant drop in performance after adding adversarial samples, while VSE-C training with contrastive adversarial samples is less vulnerable to the attacks. This phenomenon reflects that the text encoders of VSE and VSE++ do not actually make a good use of the image encodings, as the image encodings are fixed in all experiments.

Detailed attacking results are shown in Table~\ref{tab:adversarial-analysis}. The three hyper-columns show the ability of the models to defend the adversarial attack of noun, numeral, and relation-typed respectively. Among three types of attacks, VSE and VSE++ suffer least from the noun attack. As the constitution of the dataset ensures the frequency in the entire dataset of the words used for replacement, the visual grounding of these frequent nouns is easy to obtain. However, the semantics of relations (including prepositions and entity relations) or numbers are not diverse enough in the dataset, leading to the poor performance of VSE against these attacks.

\paragraph{Numeral-typed VSE-C improves the counting ability of models.} As shown in Table~\ref{tab:adversarial-analysis}, numeral-typed contrastive adversarial samples improve the counting ability of models. However, it is still not clear about where the gain comes from, as the creation of numeral-typed samples may change the form (\ie, singular or plural) of nouns to make a sentence plausible. Does the gain comes from the improved ability to distinguish singulars and plurals? 

We conduct the following evaluation to study the counting ability. We extract all the images associated with captions including plurals from our test split of MS-COCO, forming a plural-split of the dataset, and generate only plural (numeral)-typed contrastive adversarial samples (changing the numerals) \wrt the plurals in the captions. We report the performance of VSE++ and numeral-typed VSE-C on this plural split in Table~\ref{tab:plural}. It clearly shows that what VSE-C does not only distinguish singulars against plurals, but also, at least, distinguish plurals against other plurals (\eg, 3 vs. 5).

It is worth noting that such counting ability is still not evaluated completely due to the limitation of the current MS-COCO test split. We find that 99.8\% of the plurals in MS-COCO test set comes from one of ``two'', ``three'', ``four'' and ``five''. This may reduce the counting problem to a much simpler classification one. 

\begin{table}[t]
\centering
\begin{tabular}{l|rrrr|rrrr}
\hline
\textbf{Model} & \multicolumn{4}{c}{\textbf{MS-COCO Test}} & \multicolumn{4}{|c}{\textbf{MS-COCO Test (w/. adversarial) }} \\
&R@1 & R@10 &Med r.& Mean r.& R@1 &R@10 &Med r. &Mean r.\\
\hline
VSE & 47.7 & 87.8 & 2.0 & 5.8 & 28.0 & 71.6 & 4.0 & 11.7\\
VSE++ & \textbf{55.7} & \textbf{92.4} & 1.0 & \textbf{4.3} & 35.6 & 72.5 & 3.0 & 11.8\\
\hline
VSE-C (+n.) & 50.7 & 90.7 &1.0 & 5.2 & 40.3 & 80.2 & 2.0 & 9.2\\
VSE-C (+num.) & 53.3 & 90.2 & 1.0 & 5.8 & 46.9 & 86.3 & 2.0 & 6.9 \\
VSE-C (+rel.) & 52.4 & 89.0 & 1.0 & 5.7 & 42.3 & 82.5 & 2.0 & 7.2\\
VSE-C (+all) & 50.2 & 89.8 &  1.0 & 5.2 & \textbf{47.4} & \textbf{88.8} & 2.0 & \textbf{5.5}\\
\hline
\end{tabular}
\caption{\label{tab:adversarial-attack} Evaluation on image-to-caption retrieval. Although VSE++~\cite{faghri2017vse++} obtains the best performance on original MS-COCO test set, it is more vulnerable to the caption-specific adversarial attack compared with the proposed VSE-C, and so does VSE~\cite{kiros2014unifying}.}
\end{table}

\begin{table}[t]
\centering
\begin{tabular}{l|rrr|rrr|rrr}
\hline
\textbf{Model} & \multicolumn{3}{c}{\textbf{MS-COCO Test (+n.)}} & \multicolumn{3}{|c}{\textbf{MS-COCO Test (+num.) }} & \multicolumn{3}{|c}{\textbf{MS-COCO Test (+rel.)}} \\
& R@1 & R@10 & Mean r. & R@1 & R@10 & Mean r. & R@1 & R@10 & Mean r. \\
\hline
VSE & 37.6 & 85.8 & 6.9 & 38.5 & 82.3 & 7.7 & 30.7 & 76.7 & 8.8 \\
VSE++ & 45.7 & 89.1 & 5.5 & 45.9 & 82.3 & 7.2 & 42.3 & 80.0 & 7.6\\
\hline
VSE-C (+n.) & 49.2 & 88.4 & 5.7 & 42.1 & 80.3 & 9.1 & 40.4 & 83.3 & 7.1\\
VSE-C (+num.) & \textbf{51.0} & \textbf{89.5} & 6.1 & \textbf{53.3} & \textbf{90.2} & 5.8 & 49.0 & 87.0 & 6.6\\
VSE-C (+rel.) & 48.0 & 88.8 & \textbf{5.3} & 45.4 & 83.9 & 6.7 & \textbf{50.1} & \textbf{90.2} & \textbf{4.9}\\ 
VSE-C (+all.) & 49.4 & 89.3 & \textbf{5.3} & 49.9 & 89.6 & \textbf{5.2} & 47.9 & 89.4 & 5.3\\
\hline
\end{tabular}
\caption{Detailed results on each type of adversarial attack. Training VSE-C on one class gains the best performance on robustness against the adversarial attack of the class itself. In addition, training with numeral-typed adversarial samples helps improve the robustness against noun-typed and relation-typed attack. We hypothesis that this is attributed to the singularization or pluralization of the corresponding nouns in the process of numeral-typed adversarial sample generation. }
\label{tab:adversarial-analysis}
\end{table}
\begin{table}[!t]
\centering
\begin{tabular}{l|p{0.15\textwidth}p{0.15\textwidth}p{0.15\textwidth}}
\hline
\textbf{Model} & \multicolumn{3}{c}{\textbf{MS-COCO Test (plural split, +plurals)}} \\
& \multicolumn{1}{r}{~~~~~~~R@1} & \multicolumn{1}{r}{~~~~~~~~R@10} & \multicolumn{1}{r}{Mean r.~~~} \\
\hline
VSE++ & \multicolumn{1}{r}{43.7} & \multicolumn{1}{r}{78.3} & \multicolumn{1}{r}{9.1~~~}\\
VSE-C (+num.) & \multicolumn{1}{r}{\textbf{50.6}} & \multicolumn{1}{r}{\textbf{84.4}} & \multicolumn{1}{r}{\textbf{7.8}~~~} \\
\hline
\end{tabular}
\caption{\label{tab:plural} Results on plural-typed adversarial attack to the plural split of MS-COCO test set. This split consists of 205 images, together with the 1,025 original captions. VSE-C outperforms VSE++ by a large margin on all the three considered metrics.}
\end{table}

\subsection{Saliency visualization}
\begin{figure}[h]
\centering
\includegraphics[width=\textwidth]{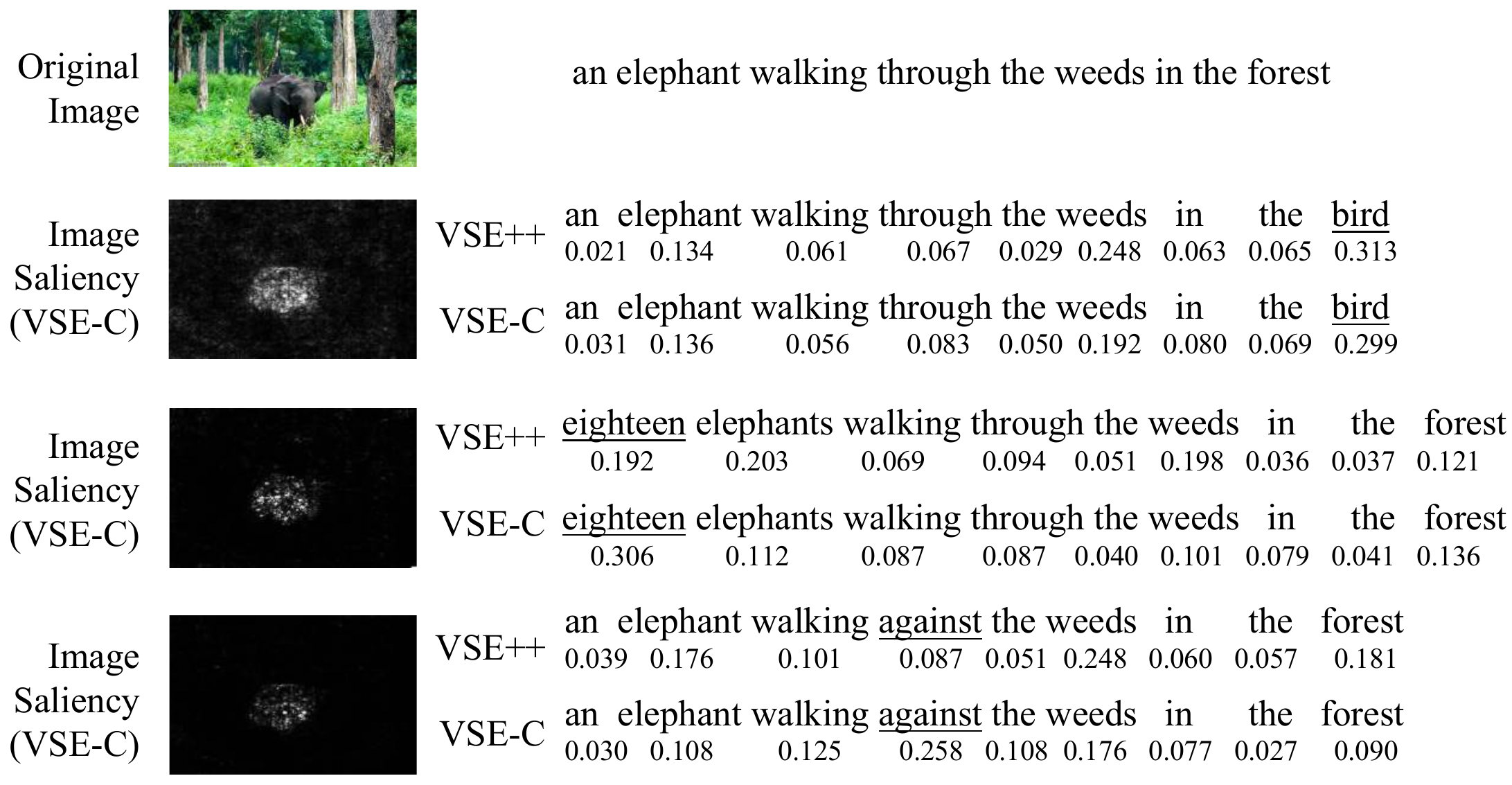}
\caption{Saliency analysis on adversarial samples. The left column shows the saliency of VSE-C on the image (what is the difference between the image and the image you imagine from caption $c'$), while the right column shows the saliency of both VSE++ and VSE-C on the caption (what is the difference between the caption and the caption you summarize from image $i$). The magnitude of values indicates the level of saliency. For better visualization, the image saliency is $L_{\inf}$-normalized and the caption saliency is $L_1$-normalized. For all the three classes of textual adversarial samples, the image encoding model (ResNet-152) almost only focuses on the main part of the image, \textit{i.e.}, elephant. For numeral-typed and relation-typed adversarial samples, VSE-C pays much more attention to the manipulated segments of the sentence than VSE++.}
\label{fig:saliency}
\vspace{-1em}
\end{figure}

Given an image-caption pair and its corresponding textual adversarial samples, we are interested in the following question: what is the semantic distance between the image an adversarial caption? In other words, \emph{which part in the image or caption, in particular, makes them semantically different?}

We visualize the saliency on input images and captions \wrt changes in sentence semantics. Specifically, given an image-caption pair $(i, c)$, we manually modify the semantics of the caption $c$ with the techniques introduced in Section~\ref{sec:method:manip}, and obtains $c' \neq c$. We compute the saliency of $i$ or $c'$ \wrt this change by visualizing the Jacobian:
\begin{equation} 
\textbf{J} = \nabla_i s(i, c') = \nabla_i W_i^T f(i; \theta_i) \cdot W_c^T g(c'; \theta_c),
\end{equation}
where $s(i, c')$ is the similarity metric for image-caption pairs.


Shown in Figure~\ref{fig:saliency}, as for captions, VSE-C captures the change in sentence semantics and thus possesses large saliency on the manipulated words. In contrast, although trained with hard-negative mining, it is difficult for VSE++ to capture differences other than nouns.

Interestingly, the saliency of images shows less correlated response to semantics changes while the replaced word is not the major component in the image. We attribute this to the image embedding extractor, ResNet, because it is pre-trained on the ImageNet classification task. As the ResNet learns to produce shift-invariant features focusing on the major components (or concepts) of images, it inevitably learns less about secondary (and other) concepts.

\subsection{Correlate words and objects}
As only textual adversarial samples are provided during the training, the model may overfit the training samples by memorizing incorrect co-occurrence of words or concepts. To quantitatively evaluate the learned word embeddings, we conduct experiments on word-level image-to-word retrieval. Specifically, we first examine how each noun is linked with a visual object. This task shows the concrete link between words and image concepts, which supports the effectiveness of adversarial samples in enforcing the learning of visually-grounded semantics beyond co-occurrence memorizing.

\paragraph{Dataset}
Based on captions, we extract \textit{positive objects} for each image in MS-COCO dataset by detecting heads of noun phrases using SpaCy. As mentioned in Section~\ref{sec:method:manip}, we only let those objects without direct hypernymy/hyponymy relation to positive objects of the image be \textit{negative objects} to avoid ambiguity. Table~\ref{table: image-object-retrieval} shows an example of the preparation of image-object dataset. 

\begin{table}[t]
\centering
\begin{tabular}{cl}
\hline
\multicolumn{1}{c}{\textbf{Image}} & \multicolumn{1}{c}{\textbf{Captions}} \\
\hline
\multirow{5}{*}{\parbox{0.22\linewidth}{\vspace{1pt}\includegraphics[height=68pt]{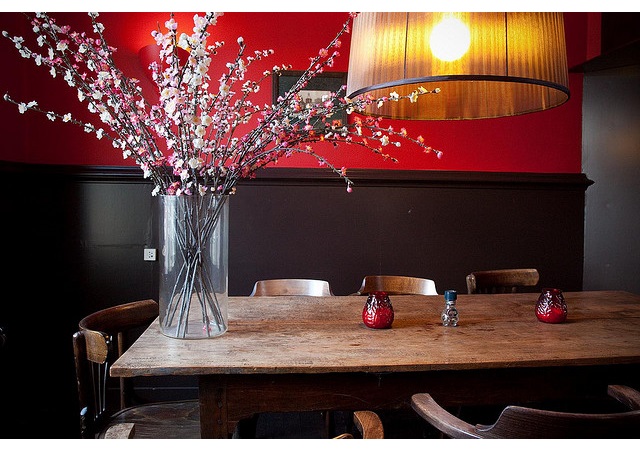}}} & A \underline{table} with a huge glass \underline{vase} and fake \underline{flowers} come out of it. \\
& A \underline{plant} in a \underline{vase} sits at the end of a \underline{table}. \\
& A \underline{vase} with \underline{flowers} in it with long \underline{stems} sitting on a \underline{table} with \underline{candles}. \\
& A large \underline{centerpiece} that is sitting on the \underline{edge} of a dining \underline{table}.\\
& \underline{Flowers} in a clear \underline{vase} sitting on a \underline{table}. \\
\hline
\multicolumn{2}{l}{\textbf{Positive Objects}:  table, plant, vase.} \\
\multicolumn{2}{l}{\textbf{Negative Objects}: screen, pickle, sandwich, toy, hill, coat, cat, etc.} \\
\hline
\end{tabular}
\caption{\label{table: image-object-retrieval} An example of the image-to-word retrieval dataset. We extract objects by detecting heads of noun phrases in captions. We only collect the ``object'' words with frequency higher than 200 in MS-COCO full dataset as available positive/negative objects for each image.}
\end{table}

\paragraph{Training}
Inspired by \newcite{gong2017natural}, we train an image-word alignment network through the interaction space, since this structure reflects the property of ``alignment'' better than just concatenating the feature vectors of word and image. 
In the training stage, the network is fed by batches of samples in the form of (image, word, label), where the label is 0 or 1, indicating whether the word is a negative or positive object of the image. 
Let $\mathbf{v}_W(w)$ denote the embedding of word $w$ and $\mathbf{v}_I(img)$ denote the feature vector of image $img$ extracted by ResNet-152~\cite{he2016deep}. 
As shown in Figure~\ref{fig: object-retrieval-network}, we use the full interaction matrix $\mathbf{v}_W(w)\mathbf{v}_I(img)^T$ as the feature for object retrieval.
While fixing both the image and word features, we only tune the parameters of multi-layer perceptron (MLP).

\begin{center}
\begin{minipage}[h]{0.58\textwidth}
\vspace{0.9em}
\centering
\includegraphics[width=\textwidth]{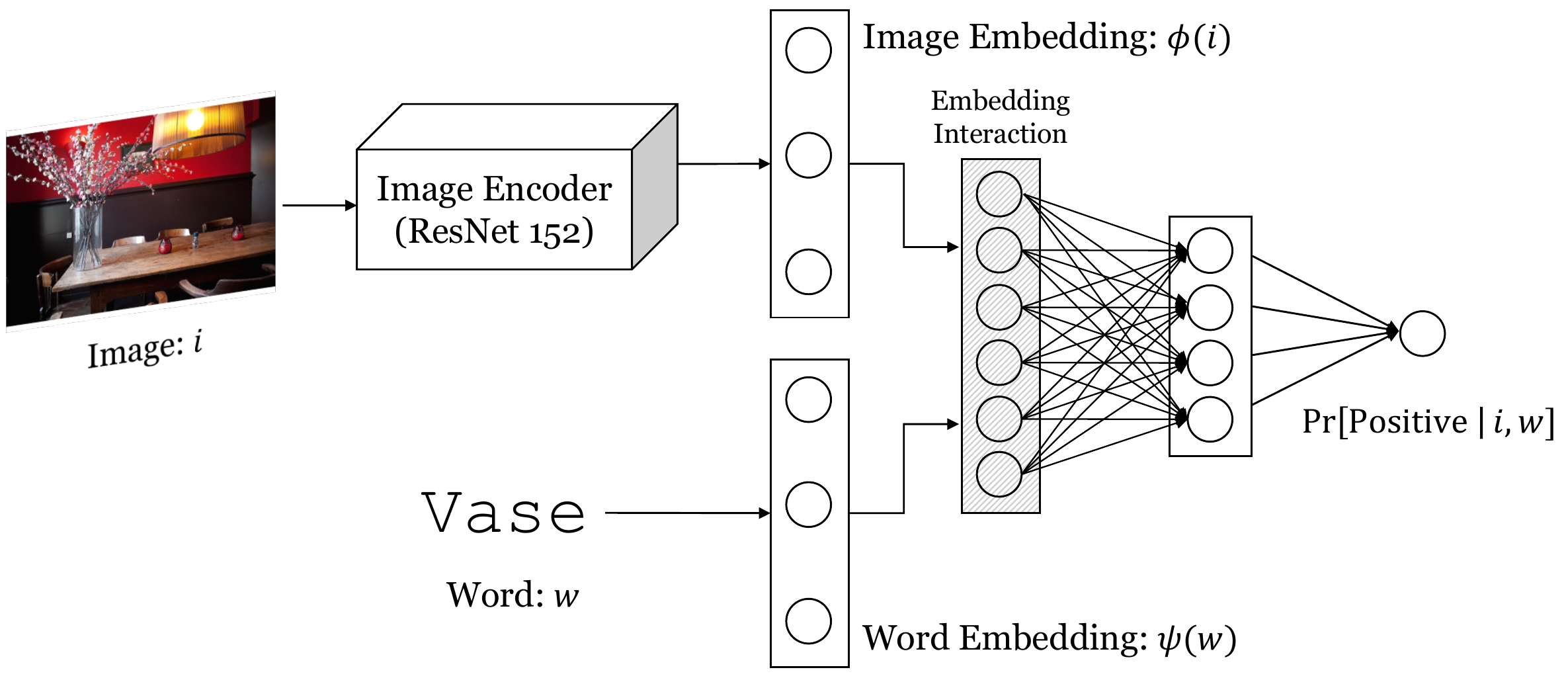}
\captionof{figure}{\label{fig: object-retrieval-network} Model structure for image-to-word retrieval. The network is trained though the interaction space. Note that only parameters in MLP are tuned during training.}
\end{minipage}
\hfill
\begin{minipage}[h]{0.38\textwidth}
\centering
\begin{tabular}{lr}
\hline
\textbf{Model} & \textbf{MAP} \\
\hline
GloVe & 58.7 \\
VSE & 61.7 \\
VSE++ & 61.1 \\
\hline
VSE-C (+all) & 62.2 \\
VSE-C (+n.) & \textbf{62.8} \\ 
VSE-C (+rel.) & 62.3 \\
VSE-C (+num.) & 62.0 \\
\hline
\end{tabular}
\captionof{table}{\label{table: result-object-retrieval} Evaluation result (MAP in percentage) on image-to-word retrieval.
}
\end{minipage}
\end{center}

\paragraph{Testing}
We use mean average precision (MAP), which is a widely-applied metric in information retrieval, to evaluate the performance of the word embeddings.
For each image, we treat it as a query. 
The average precision (AP) is defined by 
\begin{equation}
AP = \frac{\sum_{k=1}^{n}P(k) \times{positive(k)}}{\rm number~of~positive~objects}
\end{equation}
where $n$ is the quantity of objects in data base, \textit{i.e.}, both positive and negative objects, $P(k)$ is the precision at cut-off $k$ in the list, $positive(k)$ is an indicator function equaling 1 if the object at rank $k$ is a positive one, 0 otherwise~\cite{turpin2006user}. 

Based on the definition of AP, MAP can be computed by $MAP = \frac{\sum_{i=1}^{|Q|} AP(i)}{|Q|}$, where $Q$ is the query set, \textit{i.e.}, image set. It is worth noting that the database for retrieval of each query may be different from others, which is similar to Section~\ref{sec: adversarial-attack}. 

\paragraph{Results} We show the evaluation results in Table~\ref{table: result-object-retrieval}. It is as expected that VSE-C (+n.) achieves the best performance in the image-object retrieval task. All of the VSE-C models outperform the baselines produced by VSE~\cite{kiros2014unifying}, VSE++~\cite{faghri2017vse++} and GloVe~\cite{pennington2014glove}, showing the concrete link between learned word semantics and visual concepts. With surprise, VSE-C with only relation adversarial samples shows comparable performance as VSE-C with noun adversarial samples. This further supports the effectiveness of sentence-level manipulation (relation-shuffle in Figure~\ref{table: contrastive-adversarial-examples}) on strengthening the link.

\subsection{Concept to word}
We quantitatively evaluate the performance of concept-to-word retrieval performance by introducing a sentence completion task. Given an image-caption pair $(i, c)$, we manually replace concept words (nouns and relational words) with blanks. A separate model is trained to fill in the blanks.

\paragraph{Dataset and implementation details}
Based on captions, we extract nouns and relational words from captions for each image in MS-COCO dataset using SpaCy. These selected words are marked as ``concept'' representatives. During training, we randomly sample a word from the representative set, and the word is masked as a blank to be filled. Given the image and the rest of the words, the model is trained to predict the embedding of the word.

\begin{figure}[t]
\centering
\includegraphics[width=0.75\textwidth]{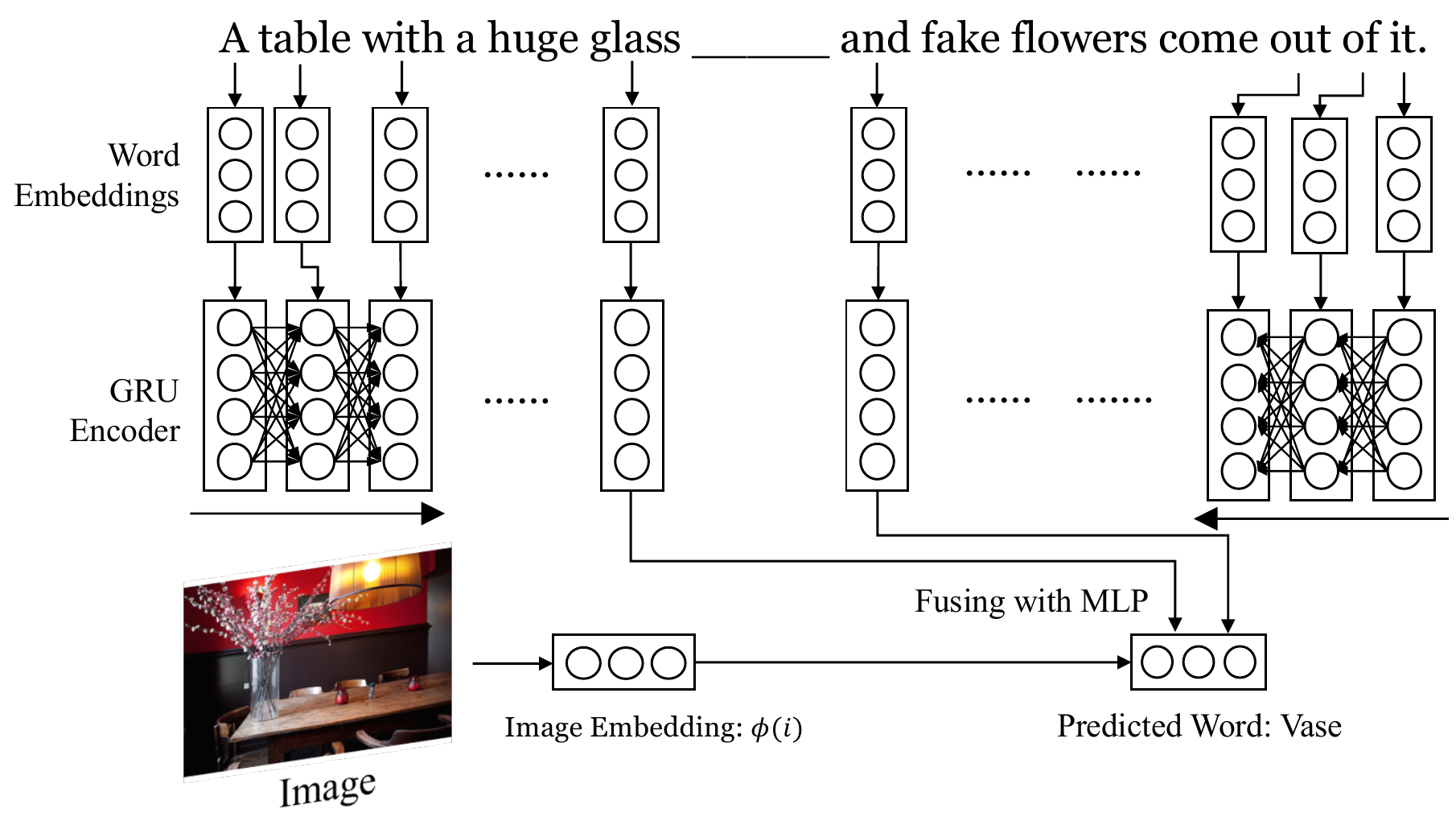}
\caption{\label{fig: fill-in-network} Model structure for fill-in-the-blank.}
\end{figure}

\paragraph{Model} The sentences with blank are encoded by two mono-directional GRU layer. The words before the blank and after the blank are separately encoded using $\text{GRU}_f$ and $\text{GRU}_b$ respectively. The image feature extracted from a pre-trained ResNet152 is then concatenated with the last output of both GRUs. The prediction of embedding is made by a two-layer MLP taking in the concatenated feature. We use cosine similarity as the loss function. Figure~\ref{fig: fill-in-network} shows the demonstration of our fill-in-the-blank model.

\paragraph{Results}
We present in Table~\ref{tab:concepttoword} the performance of the proposed VSE-C on filling in both nouns and prepositions. VSE-based models outperform GloVe without visual grounding and concretely correlate word semantics with image embeddings. We found that only small gaps exist between VSE++ and VSE-C on preposition filling, which again shows the limited diversity on visual relations within the dataset.

\begin{table}[t]
\centering
\begin{tabular}{l|rr|rr|rr}
\hline
\textbf{Model} & \multicolumn{2}{c|}{\textbf{Noun Filling}} & \multicolumn{2}{c|}{\textbf{Prep. Filling}} & \multicolumn{2}{c}{\textbf{All (n. + prep.)}} \\
& R@1 & R@10 & R@1 & R@10 & R@1 & R@10 \\
\hline
GloVe & 23.2 & 58.8 & 23.3 & 79.9 & 23.3 & 66.6 \\
VSE++ & 25.0 & 61.7 & 34.9 & 84.9 & 28.4 & 68.1 \\
VSE-C (ours) & \textbf{27.3} & \textbf{62.9} & \textbf{35.2} & \textbf{85.2} & \textbf{30.0} & \textbf{70.98}\\
\hline
\end{tabular}
\caption{\label{tab:concepttoword} Evaluation result on the fill-in-the-blank task (in percentage). The word embeddings learned by VSE-C with all classes of contrastive adversarial samples help reach a better performance than those learned by VSE++~\cite{faghri2017vse++}. }
\vspace{-1em}
\end{table}

\section{Discussion and conclusion}
In this paper, we focus on the problem of learning visually-grounded semantics using parallel image-text data. With extensive experiments on adversarial attacks against existing frameworks~\cite{kiros2014unifying,faghri2017vse++}, we obtain new insights on the limitation of datasets as well as frameworks. (1) Even for large-scale datasets such as MS-COCO captioning, the large gap between the number of possible constitutions of real-world visual semantics and the size of dataset still exists. (2) Existing models are not powerful enough to fully capture or extract the information contained in visual embeddings.

We propose VSE-C, introducing contrastive adversarial samples in the text domain and an intra-pair hard-example mining technique. To delve deeper into the embedding space and its transferability, we study a set of multi-modal tasks both qualitatively and quantitatively. Beyond being robust to adversarial attacks on image-to-caption retrieval tasks, experimental results on image-to-word retrieval and fill-in-the-blank  reveal the correlation between the learned word embeddings and visual concepts.

VSE-C also demonstrates a general framework for augmenting textual inputs considering semantical consistency. The introduction human priors and knowledge bases alleviates the sparsity and non-contiguity of languages. We hope the framework and the released data are beneficial for building more robust and data-efficient models.



\bibliographystyle{acl}
\bibliography{coling2018}

\newpage
\appendix
\section{Semantic Overlap Table of Frequent Prepositions}
\label{appendix: relation}
\begin{table}[ht]
\centering
\begin{tabular}{cl}
\hline
\textbf{Set \#} & \multicolumn{1}{c}{\textbf{Words in a Semantical Set}} \\
\hline
1& towards, toward, beyond, to \\
2& behind, after, past \\
3& outside, out \\
4& underneath, under, beneath, down, below \\
5& on, upon, up, un, atop, onto, over, above, beyond \\
6& in, within, among, at, during, into, inside, from, between \\
7& if, while \\
8& with, by, beside \\
9& around, like \\
10& to, for, of \\
11& about, within \\
12& because, as, for \\
13& as, like \\
14& near, next, beside \\
15& though \\
16& thru, through \\
17& besides, along \\
18& against, next, to \\
19& along, during, across, while \\
20& off, out \\
21& without \\
22& than \\
23& before \\
 \hline
\end{tabular}
\caption{\label{table: preps} Manually annotated semantic overlap table.}
\end{table}

Table~\ref{table: preps} shows our manually annotated semantic overlap sets. 
Prepositions in each row has overlap in semantics, \textit{i.e.}, can be replaced by each other in some level.
A preposition can appear in several sets. 

\section{Training Details of VSE-C}
In all experiments, we use Adam~\cite{kingma2014adam} as the optimizer of which the learning rate is set to 1e-3, with the batch size of 128. 
The learning rate is updated by multiplying $0.1$ after every 15 epochs.
We do not apply any regularization or dropout term.
Word embeddings are initialized with the 300-dimensional GloVe~\cite{pennington2014glove}\footnote{\url{http://nlp.stanford.edu/data/glove.840B.300d.zip}}. 
The text encoder is a bidirectional 512-dimensional (in total 1024D) 1-layer GRU. 
The dimensionality of joint (multimodal) embedding is also 1,024.
Empirically, with training data and hyper-parameters fixed, there is no significant variance in performance caused by different random seeds for the sampling.

\end{document}